\documentclass[runningheads]{llncs}
\usepackage[T1]{fontenc}
\usepackage{graphicx}
\usepackage[colorlinks=true, linkcolor=blue, urlcolor=blue, citecolor=blue]{hyperref}
\usepackage{xcolor}
\definecolor{customgreen}{HTML}{00A000} % 定义名为 customgreen 的绿色
\usepackage{tcolorbox}
\usepackage{booktabs}
\usepackage{url} 
\usepackage{tabularx}
\usepackage{authblk}
\begin{document}
\title{Keypoint-Integrated Instruction-Following Data Generation for Enhanced Human Pose and Action Understanding in Multimodal Models}
\titlerunning{Keypoint-Integrated Instruction-Following Data Generation}
% If the paper title is too long for the running head, you can set
% an abbreviated paper title here
%
\author{Dewen Zhang\inst{1}\orcidID{0009-0002-5145-0100} 
\and 
Wangpeng An\inst{2}\orcidID{0000-0002-1869-1837}
\and 
Hayaru Shouno\inst{1}\orcidID{0000-0002-2412-0184}}
\authorrunning{D. Zhang et al.}
% First names are abbreviated in the running head.
% If there are more than two authors, 'et al.' is used.
%
\institute{Department of Informatics, Graduate School of Informatics and Engineering, The University of Electro-Communications, Tokyo 182-8585, Japan\\
\email{\{zhangdewen,shouno\}@uec.ac.jp}\\
\and
TikTok Inc, 1199 Coleman Ave, San Jose, CA 95110\\
\email{anwangpeng@gmail.com}}
\maketitle              % typeset the header of the contribution
\begin{abstract}
Current vision-language multimodal models are well-adapted for general visual understanding tasks. However, they perform inadequately when handling complex visual tasks related to human poses and actions due to the lack of specialized vision-language instruction-following data. We introduce a method for generating such data by integrating human keypoints with traditional visual features such as captions and bounding boxes, enabling more precise understanding of human-centric scenes. Our approach constructs a dataset comprising 200,328 samples tailored to fine-tune models for human-centric tasks, focusing on three areas: conversation, detailed description, and complex reasoning. We establish a benchmark called Human Pose and Action Understanding Benchmark (HPAUB) to assess model performance on human pose and action understanding. We fine-tune the LLaVA-1.5-7B model using this dataset and evaluate it on the benchmark, achieving significant improvements. Experimental results show an overall improvement of 21.18\% compared to the original LLaVA-1.5-7B model. These findings highlight the effectiveness of keypoint-integrated data in enhancing multimodal models. Code is available at \url{https://github.com/Ody-trek/Keypoint-Instruction-Tuning}.

\end{abstract}
\section{Introduction}
The development of multimodal models integrating vision and language has become a central focus in artificial intelligence (AI) research~\cite{1,2,3}. Models like LLaVA (Large Language and Vision Assistant)~\cite{5} bridge the gap between visual perception and natural language understanding, utilizing large language models (LLMs) and visual encoders to process a wide range of image-related tasks~\cite{6,7,8}.

Despite their achievements, current models struggle with specialized tasks that require nuanced understanding of human activities, particularly those involving poses and actions. This limitation constrains their application in assistive robotics, healthcare, and human-computer interaction ~\cite{9,10,11,12}. A significant challenge is the lack of specialized vision-language instruction-following data. While LLaVA introduces a method for converting image-text pairs into instruction-following data using GPT-4~\cite{13}, it primarily relies on image captions and object bounding boxes, which lack the precision needed to interpret complex human activities. Consequently, models trained on such data show limited performance in tasks requiring detailed understanding of human pose and action.

To address this gap, we propose a novel approach that integrates human keypoints into the instruction-following data generation process. Our keypoint-integrated method provides a more comprehensive representation of human pose and action, enabling models to reason not just about objects in an image, but about how individuals interact with those objects and each other. This significantly enhances the model's ability to describe human movements, reason about their purposes, and respond to queries about human interactions.

Our contributions are threefold: (1) We introduce a method for generating vision-language instruction-following data by integrating human keypoints, filling a critical gap in existing multimodal models; (2) We demonstrate substantial improvements in human-centric visual tasks through comprehensive experiments comparing our fine-tuned enhanced LLaVA-1.5-7B model with its original version and other models; (3) We offer insights into how different types of fine-tuning data impact model capabilities for specific domains.

\section{Related Work}
\subsection{Instruction-Following Multimodal Models}
The LLaVA model~\cite{5} has made noteworthy progress by integrating vision encoders with LLMs to tackle various vision-language tasks. Similarly, models such as $\mathit{V}^*$~\cite{14}, Qwen2-VL~\cite{15}, VisionLLM~\cite{16} and Flamingo~\cite{17} have been developed for general vision understanding. Although effective for image description and elementary visual reasoning, these models aren't specifically designed for interpreting detailed human poses and actions. We introduce a method for generating instruction-following data specifically for human pose and action understanding by leveraging human keypoint information alongside traditional visual features. By integrating this enriched dataset into the fine-tuning process of the LLaVA-1.5-7B model, we enhance its capacity for complex reasoning and detailed description of human-centric activities.

\subsection{Multimodal Human-Centric Visual Understanding}
Traditional human activity recognition typically relies on distinct models for specific tasks~\cite{18,19,20,21}, but recent research shows a trend toward unifying these capabilities within a single multimodal framework. For instance, ChatPose~\cite{22} uses LLMs to combine language-based reasoning with visual input for understanding and generating 3D human poses. Our approach also merges visual and language processing capabilities but differs by focusing on instruction-following data specific to human pose and action scenarios. Unlike ChatPose, which employs SMPL~\cite{23} pose parameters for 3D pose representation, our work remains in a 2D context while enhancing interpretative abilities through diverse instruction-following data that encourages the model to associate human pose and action with user queries and responses.

\section{Keypoint-Integrated Visual Instruction Data Generation}
\label{section 3}
\begin{sloppypar}
Large-scale multimodal datasets like LAION-5B~\cite{24}, CC-12M~\cite{25} and COYO-700M~\cite{26} have advanced vision-language models. However, leveraging these datasets specifically for instruction-following tasks involving nuanced understanding of human pose and action remains underexplored.
\end{sloppypar}

Previous research such as LLaVA demonstrates promising results in generating visual instruction-following data using symbolic representations (captions and bounding boxes) to prompt language-only GPT-4~\cite{13}. Our approach enhances this foundation by integrating human keypoint data into the instruction-following data generation process. While LLaVA focuses primarily on captions and bounding boxes to represent visual content, our method enriches this representation by including annotations of keypoints, which capture precise positions and movements of human body parts within the scene.

\begin{figure*}[!t]
    \centering
    \includegraphics[width=\textwidth]{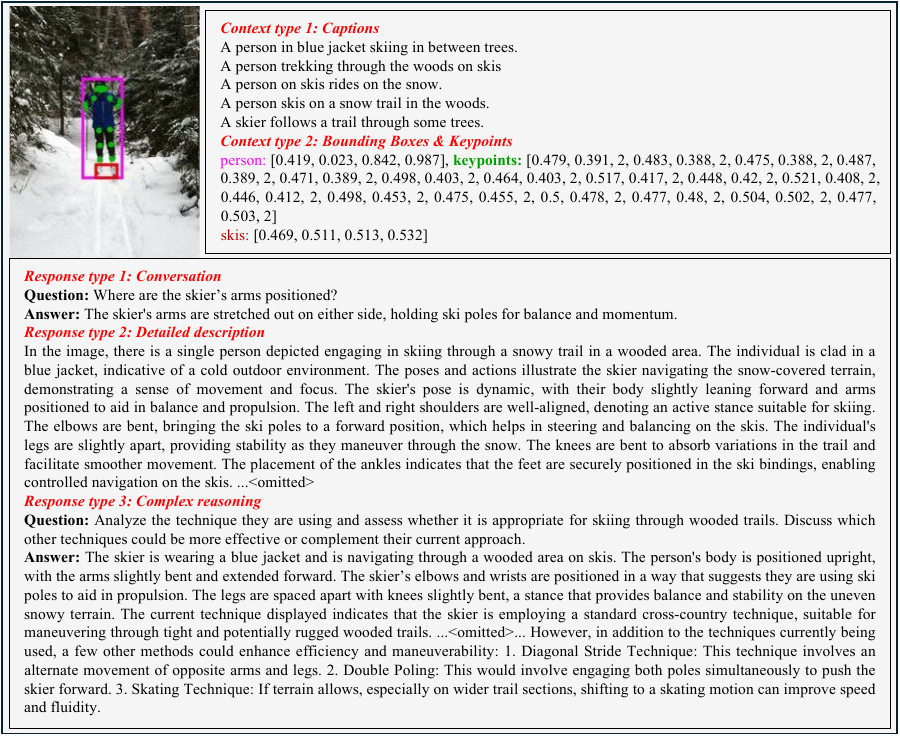} % 文件名改为您的PDF文件名
    \caption{One example to demonstrate the structure of instruction-following data. The top block displays the contexts information, including captions, bounding boxes (shown as solid rectangles in the visual image) and \textbf{keypoints} (shown as green circular markers in the visual image) used to prompt GPT-4o, and the bottom block displays the three types of responses generated. It is important to note that the visual image itself is not used to prompt GPT-4o, it is shown solely for reference purposes.}
    \label{figure 1}
\end{figure*}

To enhance the visual understanding capabilities of our model, we extend the data generation methodology originally used in LLaVA by incorporating human-centric annotations, using GPT-4o~\cite{27} as a teacher model. To represent images as visual features for prompting GPT-4o, our approach considers: (1) \textit{Captions} describing the visual scene from various perspectives; (2) \textit{Bounding boxes} localizing objects and providing spatial information; and (3) \textit{\textbf{Keypoints}} representing precise locations of joints and critical body parts. This enriched representation (example in top block of Fig.~\ref{figure 1}) allows for comprehensive understanding of human pose and action by detailing the exact positioning of body parts. The captions, bounding boxes, and keypoints annotations are obtained directly from the COCO dataset~\cite{28}.

Using these symbolic representations, we generate three distinct types of instruction-following data (example in bottom block of Fig.~\ref{figure 1}) from the COCO dataset~\cite{28}:

\begin{itemize}
\item\textbf{Conversation:} Dynamic interactions simulating real-world conversational exchanges about human poses and actions, such as asking what individuals are doing in a given scene.
\item\textbf{Detailed description:} In-depth descriptions focusing on human body language and environmental interactions, transcending simple object identification to emphasize narrative-style outputs useful in applications requiring detailed human observation.
\item\textbf{Complex reasoning:} Challenges requiring multi-step reasoning about human activities, such as understanding intentions behind specific actions or predicting next possible movements based on current poses (see detailed prompts and curation process in Table~\ref{table 1}).
\end{itemize}

\begin{sloppypar}
Due to space constraints, we present only the complex reasoning data generation approach in detail (Table~\ref{table 1}), while the approaches for conversation and detailed description follow similar methodological principles with task-specific adaptations (available in our public codebase). This approach generates 200,328 unique vision-language instruction-following samples (112,980 conversation, 45,174 detailed description, and 42,174 complex reasoning). These samples are specifically tailored to enrich the multimodal model's ability to interpret and engage with human pose and action understanding. For example, in scenarios involving skiing, as shown in Fig.~\ref{figure 1}, our approach uses keypoint data to provide nuanced understanding of the skier's posture, balance, and motion.
\end{sloppypar}

\begin{table*}[!t]
    \caption{For each query, we demonstrate the process of building the prompt for GPT-4o to gather responses, utilizing few-shot in-context learning. The examples come from annotations\_group, with each example containing an input annotation[``context'']. It is important to note that \textbf{messages} form the final prompt. In this instance, we provide the prompt used to generate the complex reasoning response for image.}
    \centering
    \begin{tcolorbox}[
        colback=gray!5, 
        colframe=black, 
        boxrule=0.3mm, 
        arc=0mm,
        sharp corners,
        boxsep=0.1mm
        ]
        \scriptsize
        \textbf{messages} = [\{\texttt{"}role\texttt{"}: \texttt{"}system\texttt{"}, \texttt{"}content\texttt{"}: f\texttt{"""}You are an AI visual assistant specializing in analyzing human poses and actions in images. You receive five sentences, each describing the same image you are observing. In addition, specific object locations within the image are given, along with detailed coordinates. These coordinates are in the form of bounding boxes and \textbf{human keypoints}, represented as (x1, y1, x2, y2) for bounding boxes and (x, y, visibility) for human keypoints, with floating numbers ranging from 0 to 1. These values correspond to the top left x, top left y, bottom right x, and bottom right y for bounding boxes, and x, y coordinates along with visibility (0: not labeled, 1: labeled but not visible, 2: labeled and visible) for human keypoints.

        The human keypoints represent the following body parts:\\
        1. nose\\
        2. left eye\\
        3. right eye\\
        4. left ear\\
        5. right ear\\
        6. left shoulder\\
        7. right shoulder\\
        8. left elbow\\
        9. right elbow\\
        10. left wrist\\
        11. right wrist\\
        12. left hip\\
        13. right hip\\
        14. left knee\\
        15. right knee\\
        16. left ankle\\
        17. right ankle\\

        The task is to use the provided caption and bounding box/human keypoint information to create a plausible question about the human poses and actions in the image, and provide the answer in detail.

        Create \textbf{complex questions} beyond describing the scene. To answer such questions, one should require first understanding the \textbf{human poses and actions}, then based on the background knowledge or reasoning, \textbf{either explain why the actions are happening that way, or provide guidance and help to the user's request}. Make the question challenging by not including the visual content details in the question so that the user needs to reason about that first.

        **Do not include any coordinates or numerical values in your explanation**. Instead, utilize the data to explain the scene using natural language. Include details like the \textbf{number of people, their actions, poses, interactions, relative positions, as well as the relationships and interactions between people and objects in the scene}. Describe how people are using objects, their proximity to objects, and any activities involving both people and objects.

        When using the information from the caption and coordinates, directly explain the scene, and do not mention that the information source is the caption or the bounding box/human keypoints. Always answer as if you are directly looking at the image.
        \texttt{"""}\}]
        \\

        \texttt{for annotation in annotations\_group:} \\
        \hspace*{2em}\textbf{messages}\texttt{.append(\{"role": "user", "content": annotation["context"]\})}

    \end{tcolorbox}
    
    \label{table 1}
\end{table*}

\section{Model Architecture and Fine-Tuning Approach}
Our enhanced LLaVA model is specifically designed to understand and interact with human-centric visual content, particularly in scenarios involving human poses and actions. This framework extends the original LLaVA architecture by fine-tuning it with keypoint-integrated instruction-following data.

\begin{figure}[h]
    \centering
    \includegraphics[width=\columnwidth]{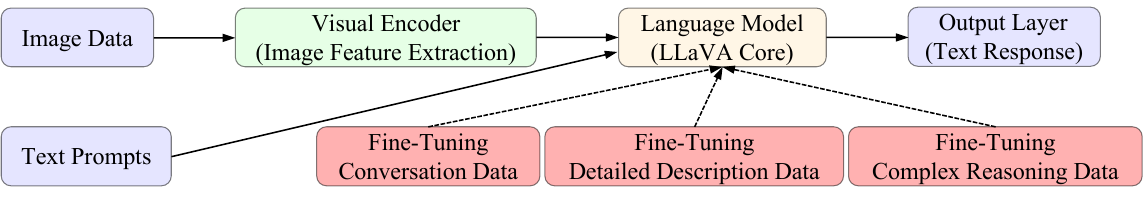} % 文件名改为您的PDF文件名
    \caption{A schematic representation of the enhanced LLaVA model architecture, showing the integration of instruction-following data for fine-tuning.}
    \label{figure 2}
\end{figure}

As shown in Fig.~\ref{figure 2}, the architecture consists of several key components:

\begin{itemize}
\item\textbf{Input Layer:} The model processes two input types: image data and text prompts. Image data, such as photographs containing humans, is processed by the visual encoder. Text prompts, comprising natural language instructions or questions, are directly fed into the language model. This bifurcated approach enables simultaneous processing of visual and linguistic modalities.

\item\textbf{Visual Encoder:} Image data is processed through a visual encoder that extracts image features. We utilize the pre-trained CLIP visual encoder~\cite{29}. The visual encoder's output is projected into a space compatible with the LLaVA core's language processing abilities.

\item\textbf{Language Model (LLaVA Core):} The LLaVA core, a language model based on the Vicuna architecture~\cite{30}, processes both the prompts and the encoded visual features, facilitating cross-modal alignment between language and vision.

\item\textbf{Fine-Tuning Modules:} We enhance the LLaVA model's capabilities by fine-tuning it with keypoint-integrated instruction-following data. This fine-tuning process utilizes the three data types discussed in Sect.~\ref{section 3}: conversation, detailed description, and complex reasoning.

\item\textbf{Output Layer:} The language model generates textual responses based on both visual and textual inputs.
\end{itemize}

With fine-tuning on keypoint-integrated instruction-following data, our model achieves strong performance in comprehending and generating responses related to human poses and actions, as demonstrated in our experimental results (Sect.~\ref{section 5}).

\section{Experiments}
\label{section 5}
We fine-tune the LLaVA-1.5-7B model to enhance its instruction-following ability in human pose and action understanding tasks, using a dataset of 200,328 unique samples generated from the COCO training dataset~\cite{28} by GPT-4o. The dataset consists of three instruction categories: conversation, detailed description, and complex reasoning (Sect.~\ref{section 3}). We train our model using the DeepSpeed framework~\cite{31} on 2×A100 GPUs, following hyperparameters similar to the original LLaVA model~\cite{5}, with adjustments for computational resources and stability, including a batch size of 8 and gradient accumulation steps of 2. Table~\ref{table 2} summarizes the hyperparameters used during fine-tuning.

\begin{table}[ht]
    \centering
    \caption{Fine-tuning hyperparameters used in our experiments.}
    \begin{tabular}{l|c} 
    \toprule
    Hyperparameter & Value \\
    \midrule
    batch size & 8 \\
    gradient accumulation steps & 2 \\
    learning rate & $2\times10^{-5}$ \\
    learning rate schedule & cosine decay \\
    learning warmup ratio & 0.03 \\
    optimizer & AdamW \\
    epoch & 1 \\
    weight decay & 0 \\
    DeepSpeed stage & 3 \\
    \bottomrule
    \end{tabular}
    \label{table 2}
\end{table}

\subsection{Qualitative Evaluation}

We conduct a qualitative evaluation to compare the responses of five models: $V^*$~\cite{14}, Qwen2-VL-7B~\cite{15}, Qwen2-VL-72B~\cite{15}, LLaVA-1.5-7B~\cite{5}, and our fine-tuned enhanced LLaVA-1.5-7B. Table~\ref{table 3} presents a detailed comparison of these models' outputs for a representative image from the COCO Validation dataset~\cite{28}, focusing on the query related to human pose and action understanding.

When asked to provide a detailed description of the poses and actions of the characters in the image, the responses from $V^*$, Qwen2-VL-7B, Qwen2-VL-72B, and LLaVA-1.5-7B models offer simple descriptions, concentrating on general aspects of the scene. Their analyses remain relatively superficial, lacking detailed explanations of the characters’ postures, movements, positions, or interactions. In contrast, our fine-tuned enhanced LLaVA-1.5-7B model delivers a more nuanced and contextually aware analysis. It identifies key elements of the main player’s posture, including specific details such as the flexion of the knees, the positioning of the elbows and wrists, and the alignment of the shoulders. Additionally, our model clearly explains how these body parts contribute to the power and precision of the player’s tennis swing. It also captures the passive involvement of two women in the scene, accurately describing their postures, attention levels, and engagement, which other models fail to recognize.

\begin{table*}[!t]
    \caption{Comparison of responses from $V^*$, Qwen2-VL-7B, Qwen2-VL-72B, LLaVA-1.5-7B, Enhanced LLaVA-1.5-7B models for the given image. \textbf{Boldfaced} content indicates key information relevant to the question.}
    \centering
    \tiny
    \resizebox{\textwidth}{!}{  % 调整表格大小以适应页面宽度
    \begin{tabular}{p{\textwidth}}  % 使用单列宽度
        \toprule
        {\textbf{A visual input example for the tennis player image}} \\
        \midrule
        \includegraphics[width=0.3\linewidth]{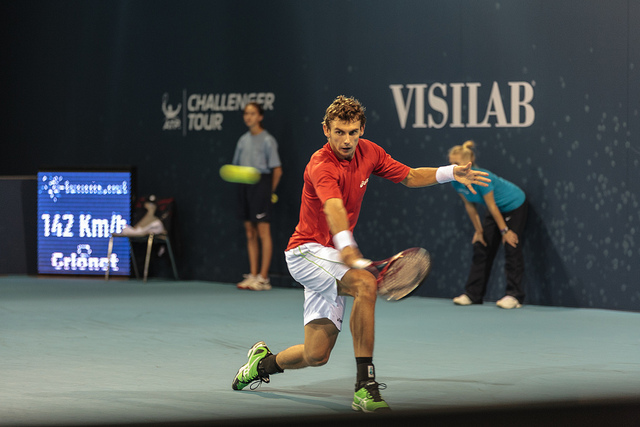} \\
        \tiny{Source:} \url{http://images.cocodataset.org/val2014/COCO_val2014_000000060992.jpg}
        \\
        \midrule
        User: Portray the poses and actions of people in the image with a rich, descriptive narrative. \\
        $V^*$: In the image, a man is swinging a tennis racket on a court, while two other people are present in the scene. One of the individuals appears to be a young girl, who might be watching the man playing tennis or waiting for her turn to participate. The man, wearing a red shirt, is actively engaged in the game, demonstrating his skills by swinging the racket with precision and power. It seems like an enjoyable and lively moment where the players share a common interest in tennis, possibly making it a family or social event where people of different ages can bond over the sport. \\
        \midrule
        User: Portray the poses and actions of people in the image with a rich, descriptive narrative. \\
        Qwen2-VL-7B: In the image, a tennis player is captured in the midst of an intense rally, showcasing his agility and skill. He is leaning forward, his body coiled like a spring, as he prepares to strike the incoming tennis ball with his racket. His eyes are focused intently on the ball, his muscles tensed and ready to spring into action. In the background, another person is holding a tennis ball, likely waiting for the right moment to serve or pass the ball to the player. This person's stance is relaxed yet attentive, as they watch the ongoing match with keen interest. A third person is standing nearby, possibly a coach or a fellow player, observing the game with a critical eye. Their presence adds an element of support and encouragement to the scene, as they watch the player's performance and offer guidance when needed. The image captures the dynamic and competitive nature of tennis, with each player fully engaged in the game and focused on their respective tasks. The vibrant colors of the court and the players' attire add to the overall energy and excitement of the scene. \\
        \midrule
        User: Portray the poses and actions of people in the image with a rich, descriptive narrative. \\
        Qwen2-VL-72B: In the image, a man is captured in the midst of an intense tennis match. He is skillfully swinging his racket with great focus and determination, aiming to strike the ball with precision and power. His body language conveys athleticism and concentration as he prepares to make contact with the ball. In the background, two other individuals can be seen observing the ongoing game. One person appears to be standing closer to the court, possibly waiting for their turn to play or simply watching the match unfold. The other individual seems to be further away, perhaps taking a break from playing or just enjoying the atmosphere of the tennis court. The scene captures the essence of a competitive yet friendly environment where people come together to engage in the sport of tennis, showcasing their skills, and sharing their passion for the game. \\
        \midrule
        User: Portray the poses and actions of people in the image with a rich, descriptive narrative. \\
        LLaVA-1.5-7B: The image captures a tennis match in progress, with a man in a red shirt and white shorts playing the game. He is in the middle of a swing, holding a tennis racket and preparing to hit the ball. Another person is standing nearby, likely watching the game or waiting for their turn to play. There are a few more people in the scene, one of them holding a tennis ball, possibly waiting for their turn to play or observing the match. The presence of multiple people in the scene suggests that this is a social event or a gathering where people are enjoying a friendly game of tennis. \\
        \midrule
        User: Portray the poses and actions of people in the image with a rich, descriptive narrative. \\
        \textbf{Enhanced LLaVA-1.5-7B:} In the image, there are three individuals engaged in a dynamic scene on a tennis court. The primary focus is on a man who is actively \textbf{swinging a tennis racket}, likely in the midst of a powerful serve or return shot. \textbf{His body is positioned with a slight forward lean, indicating the force and precision of his movement}. His \textbf{left shoulder is slightly raised}, and his \textbf{right shoulder is lower}, suggesting the follow-through of a swing. His \textbf{left elbow is bent}, and his \textbf{right elbow is extended}, with his \textbf{right wrist positioned to control the racket's direction}. His \textbf{legs are spread apart}, with his \textbf{left knee slightly bent} and his \textbf{right knee more flexed}, providing a stable base for his powerful swing. To the left of the main player, there is a woman who appears to be observing the action. She stands upright with a \textbf{relaxed posture}, her \textbf{arms hanging naturally} by her sides. Her \textbf{head is turned towards} the main player, indicating her focus on the ongoing play. Her \textbf{stance is casual}, with her \textbf{feet firmly planted} on the ground, suggesting she is not currently involved in the game but is instead watching attentively. Nearby, there is another woman who is also watching the game. She is \textbf{positioned slightly behind} the main player, with her \textbf{body turned towards} the action. Her \textbf{posture is slightly more engaged} than the observer on the left, with her \textbf{shoulders squared} and her \textbf{head turned towards} the main player. Her \textbf{arms are relaxed}, and her \textbf{legs are slightly apart}, indicating a stance that is ready to move if needed. The scene captures the intensity and focus of a tennis match, with the main player in the midst of a powerful action, while the other two individuals are observing, possibly waiting for their turn to play or simply enjoying the game. The overall atmosphere is one of concentration and athleticism, with each person contributing to the dynamic environment of the tennis court. \\
        \bottomrule
    \end{tabular}
    }
    
    \label{table 3}
\end{table*}

\begin{table}[!t]
    \caption{Ablation study on HPAUB with various training data. We prompt GPT-4o to evaluate and compare responses generated by our fine-tuned enhanced LLaVA-1.5-7B model against those from the original LLaVA-1.5-7B model. GPT-4o is tasked with assessing and providing ratings based on the overall quality of the answers, accompanied by detailed explanations.}
    \centering
    \normalsize 
    \resizebox{\textwidth}{!}{
    \begin{tabular}{l|c|c|c|c}
        \toprule
        & \textbf{Conversation} & \textbf{Detailed description} & \textbf{Complex reasoning} & \textbf{All} \\
        \midrule
        Full data & 43.67 & \textbf{67.00} & 66.67 & 59.11 \\
        Conversation & 45.00 & 32.33 & 38.67 & 38.67 \\
        Detailed description & \textbf{60.33} & 61.67 & 64.67 & \textbf{62.22} \\
        Complex reasoning & 48.00 & 55.67 & \textbf{81.00} & 61.56 \\
        LLaVA-1.5-7B~\cite{5} & 35.75 & 38.50 & 72.08 & 48.78 \\
        \bottomrule
    \end{tabular}
    }
    \label{table 4}
\end{table}

\begin{table}[!t]
    \caption{Comparative performance of $V^*$, Qwen2-VL-7B, Qwen2-VL-72B, and Enhanced LLaVA-1.5-7B on HPAUB.}
    \centering
    \small 
    % \resizebox{\columnwidth}{!}{ % 自动调整表格宽度适应列宽
    \resizebox{\textwidth}{!}{
    \begin{tabular}{l|c|c|c|c}
        \toprule
        & \textbf{Conversation} & \textbf{Detailed description} & \textbf{Complex reasoning} & \textbf{All} \\
        \midrule
        $V^*$~\cite{14} & 41.33 & 31.67 & 59.67 & 44.22 \\
        Qwen2-VL-7B~\cite{15} & 61.33 & 42.33 & 73.00 & 58.89 \\
        Qwen2-VL-72B~\cite{15} & 60.33 & 47.00 & 74.00 & 60.44 \\
        \textbf{Enhanced LLaVA-1.5-7B} & 59.22 & 58.00 & 71.00 & \textbf{62.74} \\
        \bottomrule
    \end{tabular}
    }
    \label{table 5}
\end{table}

\subsection{Quantitative Evaluation}
\begin{sloppypar}
To systematically evaluate performance, drawing inspiration from prior work~\cite{5,30}, we utilize GPT-4o~\cite{27} to measure response quality across different models. Following LLaVA's methodology, we create triplets (image, ground-truth descriptions, question) and have each model generate responses. A language-only GPT-4o then evaluates these responses on a scale of 1-10, considering helpfulness, relevance, accuracy, and detail level.
\end{sloppypar}

\subsubsection{Results.} To evaluate model performance on human-centric tasks, we construct the Human Pose and Action Understanding Benchmark (HPAUB). We randomly select 30 images containing people from the COCO Validation 2014 dataset~\cite{28} and generate three distinct types of questions for each image: conversation, detailed description, and complex reasoning, resulting in a total of 90 questions. These questions are crafted using the data generation methodology outlined in Sect.~\ref{section 3}. This benchmark evaluates the model's capability to interpret and respond to diverse human-centric visual scenarios involving human poses and actions. By systematically varying the training datasets, we analyze the impact of different types of instruction-following data on model performance, as shown in Table~\ref{table 4}. Results indicate substantial enhancements compared to the original LLaVA-1.5-7B model:

\begin{itemize}
\item
Conversation: The fine-tuned enhanced model scores 45.00 vs. original's 35.75
\item
Detailed description: 61.67 vs. 38.50
\item
Complex reasoning: 81.00 vs. 72.08
\item
Overall performance: The model fine-tuned on all three data types achieves 59.11 vs. the original LLaVA-1.5-7B's 48.78, representing a 21.18\% increase
\end{itemize}

\begin{sloppypar}
As shown in Table~\ref{table 4}, all models fine-tuned with keypoint-integrated instruction-following data outperform the baseline LLaVA-1.5-7B model that uses only captions and bounding boxes, demonstrating the significant contribution of keypoint information.
\end{sloppypar}

Using the HPAUB benchmark, we further conduct a comparative quantitative evaluation of $V^*$, Qwen2-VL-7B, Qwen2-VL-72B, and our enhanced LLaVA-1.5-7B model. The results, summarized in Table~\ref{table 5}, demonstrate that our enhanced LLaVA-1.5-7B achieves the highest overall average score of 62.74, outperforming $V^*$ (44.22), Qwen2-VL-7B (58.89), and Qwen2-VL-72B (60.44). Notably, despite having significantly fewer parameters, our 7B model surpasses the performance of Qwen2-VL-72B, which has over 10 times the parameters, highlighting the efficiency and effectiveness of our model. Compared to the second-best Qwen2-VL-72B, our model achieves an absolute improvement of 2.3 points, which translates to a relative increase of 3.8\%.

\begin{sloppypar}
\subsubsection{Discussion.} Fine-tuning the LLaVA-1.5-7B model on keypoint-integrated instruction-following data, we significantly enhance its ability to understand human pose and action. This fine-tuning process leads to substantial improvements in human-centric tasks, enabling multimodal AI systems to operate more effectively in real-world environments.
\end{sloppypar}

\section{Conclusion}
In this paper, we introduced a method for generating vision-language instruction-following data by integrating human keypoints alongside traditional captions and bounding boxes information, significantly enhancing the multimodal model's understanding of human pose and action. This innovative approach provides a more robust framework for conversations involving human activities, enabling the model to gain a deeper understanding of human-related visual contexts. Unlike the original method in LLaVA model, which only utilized bounding boxes to localize objects within images, our method leverages the detailed spatial and structural information encoded in keypoints to improve the model's interpretative and reasoning abilities. Through rigorous experimentation and evaluation, our fine-tuned models demonstrated superior performance across various tasks related to human pose and action understanding. These findings underscore the potential of integrating fine-grained human keypoint data to enhance the capabilities of multimodal AI systems. 

\subsubsection{Limitations and Future Work.} Our method focuses exclusively on static image data and thus lacks temporal modeling capability, limiting the model's understanding of dynamic human actions and interactions. Recognizing and reasoning about the temporal progression of movements is crucial in scenarios such as video analysis and sports performance evaluation. To address this limitation, future research could integrate temporal modeling techniques, such as recurrent neural networks (RNNs) or Transformers with temporal attention mechanisms, leveraging sequential data from video frames. Incorporating temporal information explicitly could enable multimodal models to better track and analyze pose dynamics over time, potentially enhancing their interpretative capabilities for real-world human behaviors in dynamic environments.

\subsubsection{Acknowledgements} This work was supported by JST SPRING, Grant Number JPMJSP2131.

%
% ---- Bibliography ----
%
% BibTeX users should specify bibliography style 'splncs04'.
% References will then be sorted and formatted in the correct style.
%
\bibliographystyle{splncs04}
\bibliography{mybibliography}

\end{document}